\documentclass{article} %
\usepackage{iclr2026_conference,times}

\usepackage{amsmath,amsfonts,bm}

\def\eqref#1{equation~\ref{#1}}

\def\1{\bm{1}}

\DeclareMathAlphabet{\mathsfit}{\encodingdefault}{\sfdefault}{m}{sl}
\SetMathAlphabet{\mathsfit}{bold}{\encodingdefault}{\sfdefault}{bx}{n}

\usepackage[table]{xcolor}
\usepackage[utf8]{inputenc} %
\usepackage[T1]{fontenc}    %
\usepackage{natbib}
\usepackage{hyperref}
\usepackage{url}
\usepackage{tcolorbox}
\usepackage{amsmath}
\usepackage{graphicx}
\usepackage{multirow}
\usepackage{colortbl}
\usepackage{booktabs}
\usepackage[linesnumbered,ruled,vlined]{algorithm2e}
\usepackage{makecell}
\usepackage{hyperref}
\usepackage{url}
\SetNlSty{textbf}{}{}  
\usepackage{wrapfig}%
\SetAlgoNlRelativeSize{-1}       %
\usepackage{subcaption}

\definecolor{citecolor}{HTML}{0071BC}
\definecolor{linkcolor}{HTML}{D32F2F}%
\definecolor{cellcolor}{HTML}{E3F2FD}
\definecolor{red}{HTML}{D32F2F}
\definecolor{magenta}{HTML}{D81B60}
\newcommand*\samethanks[1][\value{footnote}]{%
  \footnotemark[#1]
}
\hypersetup{
  colorlinks=true,
  linkcolor=linkcolor,  
  citecolor=citecolor,
  urlcolor=magenta        
}
\renewcommand{\cite}{\citep}

\title{Improved Iterative Refinement for Chart-to-Code Generation via Structured Instruction}

\author{%
  \quad \textbf{Chengzhi Xu}$^{1,}$\thanks{Equal contribution. This work was conducted at MIFA Lab (members from SJTU \& SII).}
  \quad
  \textbf{Yuyang Wang} $^{1,}$\samethanks[1]
  \quad
  \textbf{Lai Wei} $^{1,}$\samethanks[1]
  \quad
  \textbf{Lichao Sun}$^{3}$
  \quad
  \textbf{Weiran Huang}$^{1,2,}$\thanks{Correspondence to Weiran Huang (weiran.huang@outlook.com)}\\[0.3cm]
  \quad \quad \quad \quad \quad \quad \quad$^1$ School of Computer Science, Shanghai Jiao Tong University\\[0.1cm]
  \quad \quad \quad \quad \quad \quad \quad \quad $^2$ Shanghai Innovation Institute
   \quad \quad$^3$ Lehigh University
}

\iclrfinalcopy %
\begin{document}

\maketitle

\begin{abstract}
Recently, multimodal large language models (MLLMs) have attracted increasing research attention due to their powerful visual understanding capabilities. While they have achieved impressive results on various vision tasks, their performance on chart-to-code generation remains suboptimal. This task requires MLLMs to generate executable code that can reproduce a given chart, demanding not only precise visual understanding but also accurate translation of visual elements into structured code. Directly prompting MLLMs to perform this complex task often yields unsatisfactory results. To address this challenge, we propose {ChartIR}, an iterative refinement method based on structured instruction. First, we distinguish two tasks: visual understanding and code translation. To accomplish the visual understanding component, we design two types of structured instructions: description and difference. The description instruction captures the visual elements of the reference chart, while the difference instruction characterizes the discrepancies between the reference chart and the generated chart. These instructions effectively transform visual features into language representations, thereby facilitating the subsequent code translation process. Second, we decompose the overall chart generation pipeline into two stages: initial code generation and iterative refinement, enabling progressive enhancement of the final output. Experimental results show that, compared to other methods, our method achieves superior performance on both the open-source model Qwen2-VL and the closed-source model GPT-4o.
\end{abstract}

\section{Introduction}
Recently, Multimodal Large Language Models (MLLMs)~\cite{bai2023qwenvlversatilevisionlanguagemodel, lu2024deepseekvlrealworldvisionlanguageunderstanding, geminiteam2024geminifamilyhighlycapable, openai2024gpt4ocard, liu2023visualinstructiontuning}, such as GPT-4V, LLaVA, and Qwen-VL, have attracted significant attention due to their impressive capabilities in visual understanding and reasoning. Among the many downstream tasks involving multimodal inputs, chart reasoning has emerged as an important domain~\cite{xia2025chartxchartvlmversatile, masry2022chartqabenchmarkquestionanswering, han2023chartllamamultimodalllmchart, liu2024mmcadvancingmultimodalchart, he2024distillvisualchartreasoning}, as charts play a crucial role in presenting structured data in a visual format. However, recent  studies show that current MLLMs still struggle to handle a range of chart-related tasks effectively, including summarization, captioning, question answering, and particularly chart-to-code generation. The unique difficulties of chart inputs, such as their high information density and complex layout, render direct learning from vision-language corpora insufficient. Therefore, existing MLLMs generally perform poorly on chart-to-code generation benchmarks, due to their inability to accurately capture and translate multi-dimensional chart information.

To address the limitations of direct generation, some recent efforts have attempted to enhance the chart-to-code capability of MLLMs through more structured approaches. {ChartCoder}~\cite{zhao2025chartcoderadvancingmultimodallarge} improves performance by decomposing the generation process into step-by-step procedures and applying instruction tuning to open-source models. However, due to its reliance on training, ChartCoder cannot be applied to commercial closed-source models like GPT-4o, and its performance remains suboptimal. On the other hand, {METAL}~\cite{li2025metalmultiagentframeworkchart} adopts a training-free, multi-agent system that enables compatibility with both open-source and commercial MLLMs. It iteratively refines the generated code through a pipeline of generation, critique, and revision based on visual metrics. Nevertheless, METAL suffers from critical limitations: it updates code based only on the lowest-performing single metric, which often leads to unbalanced optimization, and some metrics such as color are rarely used in practice (see Table~\ref{tab:metrics_nums}). Moreover, optimizing one metric may degrade others, and relying on a single score for update decisions can result in code that is worse overall. These issues hinder METAL's ability to generate high-quality, visually faithful chart code.

\begin{wraptable}{r}{0.45\textwidth}
   \vspace{-1.0em}
  \centering
  \small  %
  \caption{Total usage counts of three metrics}
  \vspace{2mm}
  \label{tab:metrics_nums}
  \begin{tabular}{lcc}
    \toprule
    \textbf{Metric} & \textbf{Plot2Code} & \textbf{ChartMimic} \\
    \midrule
    Text    & 93  & 694 \\
    Color   & 0   & 4   \\
    Overall & 108 & 554 \\
    \midrule
    Total samples & 201&1252\\
    \bottomrule
  \end{tabular}
  
\end{wraptable}

To better understand the unique challenges behind chart-to-code generation, we conduct a detailed analysis and identify two core bottlenecks: \emph{visual understanding} and \emph{code translation}. Visual understanding refers to the MLLM's ability to extract accurate and multi-dimensional information, such as structure, text, colors, and layout, from the input chart. Code translation involves converting this rich visual information into executable and semantically faithful code. Motivated by this observation, we propose {ChartIR}, a training-free iterative framework that explicitly addresses both challenges. To enhance the visual understanding phase, we introduce a structured, multi-dimensional {chart description} mechanism that provides detailed annotations over several key visual components. To improve code translation, we design an iterative generation scheme that leverages both the original image and the chart description to guide code refinement. In each iteration, the model identifies and compares the differences between the generated chart and the reference chart across all key dimensions, including text, color, layout, and type, enabling more holistic and targeted updates.

To validate the effectiveness of ChartIR, we conduct extensive experiments on two widely-used chart-to-code benchmarks: {Plot2Code}~\cite{wu2024plot2codecomprehensivebenchmarkevaluating} and {ChartMimic}~\cite{yang2025chartmimicevaluatinglmmscrossmodal}. We evaluate our method on both open-source (Qwen2-VL) and closed-source (GPT-4o) models. Results show that ChartIR significantly outperforms both direct generation and METAL across a comprehensive set of evaluation metrics, including GPT-4o Score, low-level metrics (Text, Layout, Type, Color), and traditional image similarity metrics (SSIM, CLIP, MSE, PSNR). Notably, the improvements in GPT-4o Score, an evaluation metric closely aligned with human judgment, demonstrate that our approach leads to substantial gains in visual and structural fidelity of the generated charts. We further conduct ablation studies to isolate the effects of chart description and iterative difference-based refinement. Results show that removing either component leads to considerable performance degradation, confirming the necessity and effectiveness of each module in the ChartIR framework.

In summary, our contributions are threefold: (1) We propose {ChartIR}, a training-free, model-agnostic framework for enhancing chart-to-code generation in MLLMs; (2) We introduce a structured {chart description} mechanism that captures multi-dimensional visual features of the reference chart, which is shown to be essential for accurate code generation; and (3) We design an iterative {difference-based refinement} process that improves code quality across all visual aspects. Our method achieves state-of-the-art results on multiple benchmarks both on open-source and closed-source models and provides a robust blueprint for future research in chart-to-code generation.

\section{Related work}
\paragraph{Multimodal Large Language Models.}
Multimodal large language models (MLLMs) extend the capabilities of large language models (LLMs) by incorporating vision inputs, enabling them to perform tasks involving both text and images. Open-source MLLMs such as Qwen-VL~\cite{bai2023qwenvlversatilevisionlanguagemodel}, DeepSeek-VL~\cite{lu2024deepseekvlrealworldvisionlanguageunderstanding}, LLaVA~\cite{liu2023visualinstructiontuning}, and Mini-Gemini~\cite{geminiteam2024geminifamilyhighlycapable} have attracted attention due to their transparency, accessibility, and potential for fine-tuning. These models are often built by aligning pre-trained vision encoders with language models using multimodal instruction tuning. On the other hand, closed-source commercial models such as GPT-4o~\cite{openai2024gpt4ocard} and Claude~\cite{anthropic2024claude3} integrate proprietary vision-language pipelines and demonstrate superior performance across a wide range of multimodal tasks. In chart-related tasks, such as summarization, captioning, question answering and chart-to-code generation, these closed-source models consistently outperform their open-source counterparts. However, their closed nature restricts research access, model interpretability, and downstream adaptability, especially for tasks requiring task-specific customization or iterative optimization.

\paragraph{Chart-to-Code Generation.}
Chart-to-code generation aims to convert visual charts into executable code that reproduces the original figure. Recent studies~\cite{ xia2025chartxchartvlmversatile, han2023chartllamamultimodalllmchart, liu2024mmcadvancingmultimodalchart, zhang2025enhancingcharttocodegenerationmultimodal} have evaluated MLLMs on this task. These works show that both open-source models (e.g., Qwen-VL, DeepSeek-VL) and closed-source models (e.g., GPT-4o, Claude) still struggle with accurately reproducing charts. Among the most recent and advanced approaches, \textbf{ChartCoder} improves performance by decomposing the task into sequential substeps and training dedicated modules, but is limited to open-source models and still underperforms closed-source ones. \textbf{METAL} introduces a training-free multi-agent framework that critiques and revises code iteratively, allowing compatibility with both model types. However, it relies heavily on a single evaluation metric per iteration and lacks a unified optimization mechanism across all visual dimensions, leading to suboptimal convergence. Our method shares with METAL the idea of iterative refinement but differs by introducing a multi-dimensional description and holistic difference-based generation strategy that ensures all critical aspects of the chart are considered in every iteration.

\section{Methodology}

\begin{figure}[tp]
    \centering
    \includegraphics[width=0.95\textwidth]{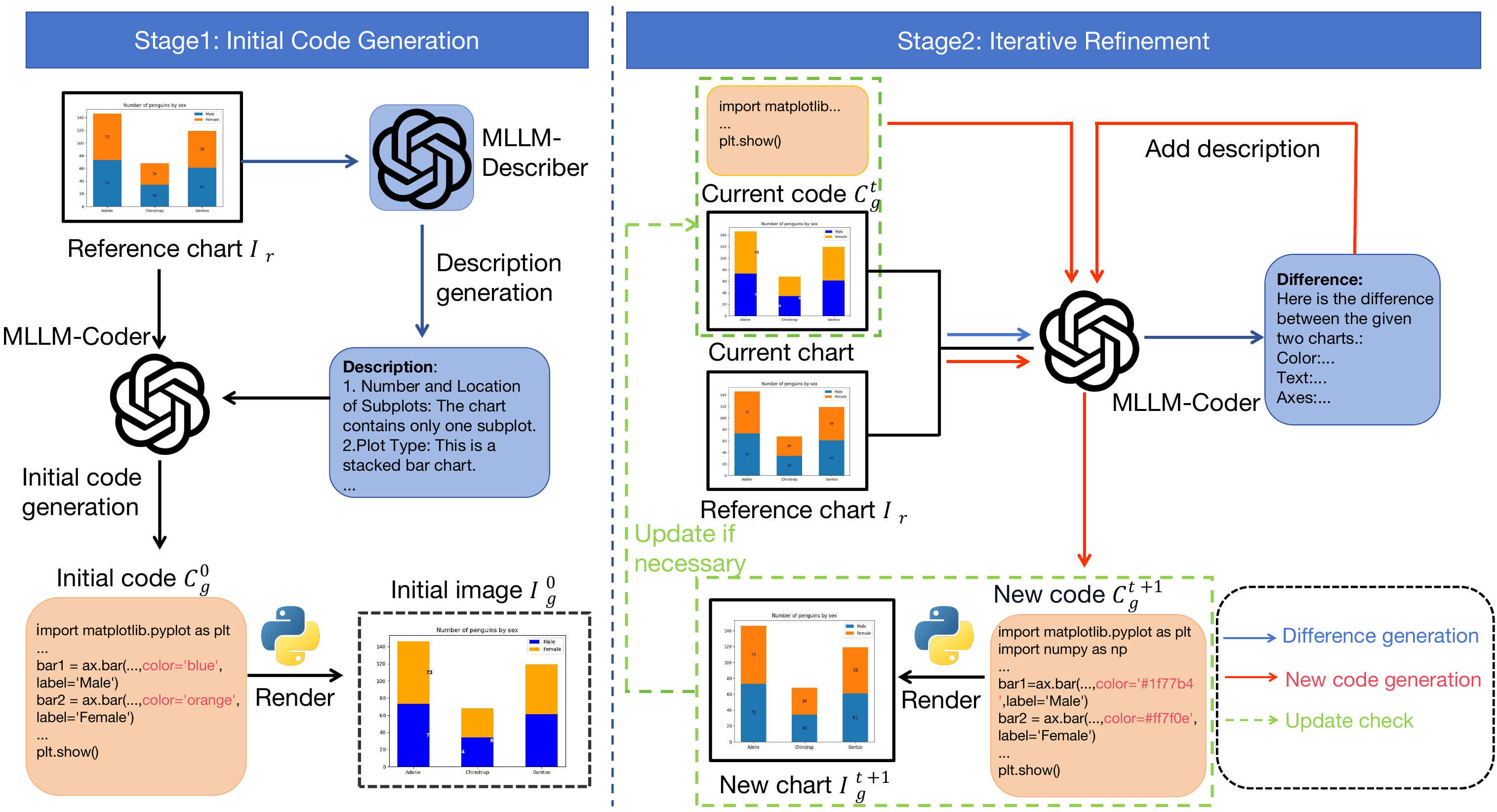} 
    \caption{Overview of the ChartIR. The method consists of two steps: 1) Initial Code Generation: First, a description is generated based on the reference chart. Then, using both the description and the reference chart, an initial code is produced. 2) Iterative Refinement: In this stage, first comparing the chart generated from the initial code with the reference chart and produces a description of the difference. Based on this difference, the reference chart, and the description of the reference chart, a new code is generated and then get a new chart. This process is repeated iteratively. The iteration continues until convergence is achieved.}
    \label{fig:flowchart}
\end{figure}

We first provide a formal problem formulation of chart-to-code generation. Given a reference chart $I_r$, we aim to generate executable code $C_g$
using an MLLM such that the resulting rendered chart minimizes visual discrepancy with the reference chart $I_r$, i.e.,
\begin{equation}
    \min_{C_g} \mathrm{disc}(\mathrm{Render}(C_g), I_r),
    \label{eq:task_definition}
\end{equation}
where $\mathrm{disc}$ represents the visual discrepancy.
Common metrics for $\mathrm{disc}$ include SSIM~\cite{2004Image}, CLIP scores~\cite{radford2021learningtransferablevisualmodels}, etc.

According to the problem definition, a natural strategy is to iteratively optimize the generated code based on the discrepancy evaluation function $\mathrm{disc}$. Figure \ref{fig:flowchart} illustrates the entire pipeline of our code generation and optimization process. Overall, our method first generates a textual description of the reference chart \(I_r\), and then produces the initial code \(C_g\) based on both the description and the reference chart \(I_r\). We then iteratively refine the code \(C_g^{new}\) at each step \(t\) by minimizing the visual discrepancy \(\mathrm{disc}(I_g^{new}, I_r)\), where \(I_g^{new} = \mathrm{Render}(C_g^{new})\). This refinement process continues until the MLLM can no longer make meaningful improvements to the generated code. In the remainder of this section, we provide a detailed description of our proposed method.

\subsection*{Stage 1: Initial Code Generation}

In the first stage, we carefully prompt the multimodal large language models (MLLMs) to produce the initial chart-generating code $C_g$ by effectively leveraging both the reference chart image $I_r$ and a detailed textual description $\delta$ of the chart.

\textbf{Description Generation.} To enhance the model’s understanding of the visual chart content, we first generate a description $\delta$ that captures key semantic and structural information from $I_r$. This description serves as an auxiliary input to guide the model during code generation. For closed-source models such as GPT-4o, which possess strong chart understanding capabilities, we directly prompt the model with the reference chart to obtain the description $\delta$ prior to code generation. In contrast, open-source models typically demonstrate weaker performance in chart interpretation. To address this limitation, we fine-tune Qwen2.5-VL~\cite{bai2025qwen25vltechnicalreport} specifically for the chart description task. We construct a training set using the ChartX~\cite{xia2025chartxchartvlmversatile} dataset by prompting GPT-4o with chart images and corresponding code to generate descriptive captions. This results in a dataset of approximately 5,000 (chart, description) pairs. We then fine-tune Qwen2.5-VL using this dataset to enable it to generate high-quality chart descriptions. 

\textbf{Code Generation.} Once the description $\delta$ is obtained, it is combined with the chart image $I_r$ and fed into the MLLM to produce the initial code $C_g$. In the ablation experiments, we demonstrate the effectiveness of incorporating chart descriptions for both open-source and closed-source models. The overall procedure of Stage 1 is summarized in {Algorithm~\ref{alg:stage1}}.

\begin{algorithm}[t]
\caption{Initial Code Generation}
\label{alg:stage1}
\KwIn{Reference chart image $I_r$, MLLM-Describer, MLLM-Coder, Render.}
Generate chart description: $\delta \leftarrow \text{MLLM-Describer}(I_r)$\;
Generate initial code: $C_g \leftarrow \text{MLLM-Coder}(I_r, \delta)$\;
Render initial image:$I_g \leftarrow \text{Render}(C_g)$\;
\Return $C_g, I_g$
\end{algorithm}

\subsection*{Stage 2: Iterative Refinement}
In the second stage, we iteratively refine the initial code $C_g$ to improve the quality of the generated chart. The generated chart image $I_g$ is first obtained by executing $C_g$. We then prompt the MLLM to describe the differences $\mu$ between $I_r$ and $I_g$. Based on this difference description, the model analyzes $I_r$, $I_g$, and $C_g$ to produce a revised version of the code, denoted as $C_{g}^{new}$. This updated code is executed to generate a new chart image $I_{g}^{new}$. Unlike METAL, which refines the code based on improvements in a specific metric at each iteration, our method allows the MLLM to consider all visual and structural aspects of the chart. The full iterative process is outlined in {Algorithm~\ref{alg:stage2}}.

\textbf{Update Criterion.} The decision to update $C_g$ to $C_{g}^{new}$ is based on the discrepancy evaluation function $\mathcal{D}$, defined in Equation~\ref{eq:task_definition}. The update occurs only when the new image $I_{g}^{new}$ is quantitatively closer to the reference image $I_r$ than $I_g$ is; that is, when $\mathrm{disc}(I_{g}^{new}, I_r) < \mathrm{disc}(I_g, I_r)$. To ensure robust and comprehensive evaluation, we aggregate multiple visual similarity metrics, including CLIP Score, DINO, SSIM, PSNR, and Hamming Distance. These are combined via a weighted average to compute the overall discrepancy score. An update is accepted only when the aggregated score for $I_{g}^{new}$ is less than that of $I_g$, ensuring improvements across multiple dimensions.

\textbf{Convergence Criterion.} To ensure convergence, we introduce a convergence counter $k$ initialized to zero. Each time the discrepancy score declines (i.e., $I_{g}^{new}$ is closer to $I_r$ than $I_g$), the counter $k$ is reset. If an iteration fails to decrease the discrepancy score, $k$ is incremented. Once the counter reaches a predefined threshold, the refinement process stops, and the final code is returned.

\textbf{Debug.} During the entire iterative refinement process, the MLLMs may occasionally generate code that fails to execute. To address this potential issue, we invoke a specialized LLM designed for code-related tasks~\cite{hui2024qwen25codertechnicalreport} to perform automatic bug fixing in real time. Specifically, we provide the model with both the detailed error message and the faulty code as input, enabling it to generate a corrected version of the generated code.

\begin{algorithm}[t]
\SetAlgoLined
\caption{Iterative Refinement}
\label{alg:stage2}
\KwIn{Reference chart image $I_r$, initial code $C_g$, description $\delta$, MLLM-Coder, Render.}
\KwOut{Final refined code.}

Initialize convergence counter: $k\leftarrow 0$\;

\Repeat{$k$ reaches a pre-determined threshold }{
    Render chart: $I_g \leftarrow \text{Render}(C_g)$\;
    Generate difference description: $\mu \leftarrow \text{MLLM-Coder}(I_r, I_g)$\;
    Refine code: $C_{g}^{new}\leftarrow \text{MLLM-Coder}(I_r, I_g, C_g, \delta, \mu)$\;
    Render new chart: $I_{g}^{new} \leftarrow \text{Render}(C_{g}^{new})$\;

    \eIf{$\mathrm{disc}(I_{g}^{new}, I_r) < \mathrm{disc}(I_g, I_r)$}{
        Refine code: $C_g \leftarrow C_{g}^{new}$\;
        Reset convergence counter: $k \leftarrow 0$\;
    }
    {
        Increment convergence counter: $k \leftarrow k + 1$\;
    }
}
\Return $C_g$
\end{algorithm}

\section{Experiments}
\subsection{Experiment setting}
The primary goal of our experiments is to evaluate whether ChartIR can enhance the ability of base multimodal models to convert chart images into code, and whether it can outperform existing approaches such as {METAL}~\cite{li2025metalmultiagentframeworkchart}. Based on our experimental setting, we conduct experiments by applying ChartIR to GPT-4o and Qwen2-VL across two benchmark datasets: {Plot2Code}~\cite{wu2024plot2codecomprehensivebenchmarkevaluating} and {ChartMimic}~\cite{yang2025chartmimicevaluatinglmmscrossmodal}. We also evaluate several baselines, including Direct Generation, and METAL. 
\paragraph{Baselines.} 
We compare our approach with the following two methods: 1) {Direct Generation:} In this setting, we directly prompt MLLMs to generate code based on the reference chart. The MLLMs receive only a generation prompt along with the reference chart, without access to any additional information or further optimization processes. 2) {METAL:} METAL employs an iterative multi-agent optimization framework to achieve the generation process.

\paragraph{Dataset.} In this work, we adopt two widely-used open-source datasets specifically designed for chart-to-code tasks: Plot2Code and ChartMimic. Plot2Code contains 132 manually curated test examples, whose plots exhibit a wide range of diversity in terms of size, textual elements, color schemes, and chart types \cite{wu2024plot2codecomprehensivebenchmarkevaluating}. It supports two evaluation settings: Direct Asking and Conditional Asking. ChartMimic consists of 500 manually filtered (figure, code, instruction) triplets for the Direct Mimic task and another 500 triplets for the Customized Mimic task \cite{yang2025chartmimicevaluatinglmmscrossmodal}. The plots in ChartMimic cover 18 regular types and 4 advanced types, demonstrating substantial diversity and complexity. In this work, we focus exclusively on the Direct Asking setting for Plot2Code and the 500 triplets for the Direct Mimic task for ChartMimic.

\paragraph{Base Model.} Our experiments are conducted under two distinct settings: closed-source and open-source. In the closed-source setting, commercial close-source models are allowed, whereas in the open-source setting, such models are excluded. The open-source setting ensures that all user data remains local, thereby preserving user privacy. Specifically, for the closed-source setting, we employ GPT-4o~\cite{openai2024gpt4ocard} as our experimental model. Given GPT-4o’s strong inherent capabilities in chart understanding, we utilize it directly to generate chart descriptions. In contrast, for the open-source setting, we adopt Qwen2-VL~\cite{wang2024qwen2vlenhancingvisionlanguagemodels} as our base model and generate chart descriptions using our fine-tuned version of Qwen2.5-VL~\cite{bai2025qwen25vltechnicalreport}.

\paragraph{Fine-Tuning.}
To further enhance the base model's chart-to-code capability, we perform instruction tuning on Qwen2.5-VL using a curated dataset derived from the ChartX~\cite{xia2025chartxchartvlmversatile} collection. Specifically, we select 5,000 chart samples, each containing an image and its corresponding Python source code. To construct high-quality multi-dimensional descriptions for each chart, we prompt GPT-4o to analyze the image and produce structured explanations based on the underlying code. The resulting dataset consists of (image, description, code) triplets, which we use to fine-tune Qwen2.5-VL. This fine-tuning step equips the model with the ability to understand and reason over visual elements in a more structured and interpretable manner, forming a stronger foundation for our iterative chart-to-code generation framework.

\paragraph{Evaluation Metric.} We employ the following metrics to evaluate the quality of the final generated charts: 1) GPT-4o Score. GPT has been widely adopted for evaluating a variety of natural language and vision tasks~\cite{yan2025gptimgevalcomprehensivebenchmarkdiagnosing,lee2024applyinglargelanguagemodels}. Therefore, we follow the evaluation benchmark Plot2Code~\cite{wu2024plot2codecomprehensivebenchmarkevaluating} and adopt the same GPT-4o prompt: we feed both the ground-truth chart and the generated chart into GPT-4o, then guide it to output a quality score ranging from 0 to 10. To ensure the reliability and precision of the GPT-4o score, we obtain three independent GPT-4o scores and compute their mean as the final score. 2) Four low-level metrics. Unlike GPT-4o Score which serves as high-level metrics, We follow the four low-level evaluation metrics proposed in the ChartMimic~\cite{yang2025chartmimicevaluatinglmmscrossmodal}: Text Score, Layout Score, Type Score and Color Score. They are designed to evaluate four key low-level visual components of a chart: Text, Layout, Type and Color. 3) Some other traditional metrics such as PSNR, SSIM and MSE \cite{doi:10.1049/el:20080522, 2004Image} which are widely used for assessing image similarity.
\subsection{Experiment Results}

\begin{table}[t]
\centering
\caption{Evaluation results on the Plot2Code and ChartMimic dataset. ChartIR outperforms both Direct Generation and METAL under most evaluation metrics for both open-source (Qwen2-VL) and closed-source (GPT-4o) models.}
\label{tab:results}
\renewcommand{\arraystretch}{1.3}
\setlength{\tabcolsep}{6pt}
\resizebox{0.95\linewidth}{!}{
\begin{tabular}{cllccccccc|c}
\toprule
\multirow{2}{*}{\textbf{Dataset}} &\multirow{2}{*}{\textbf{Base Model}} & \multirow{2}{*}{\textbf{Method}} & \multicolumn{4}{c}{\textbf{Low-Level Metrics}} & \multicolumn{3}{c|}{\textbf{Traditional Metrics}} & \multirow{2}{*}{\textbf{GPT-4o Score}} \\
\cmidrule(lr){4-7} \cmidrule(lr){8-10}
& & & \textbf{Text} & \textbf{Type} & \textbf{Layout} & \textbf{Color} & \textbf{PSNR} & \textbf{SSIM} & \textbf{MSE} & \\
\midrule
\multirow{6}{*}{\rotatebox{90}{{Plot2Code}}}&
\multirow{3}{*}{Qwen2-VL} 
& Direct   & 0.30 & 0.65 & 0.60 & 0.45 & 11.91 & 0.62 & 16958 & 3.12 \\
& & METAL    & 0.47 & 0.63 & 0.75 & 0.50 & 12.00 & 0.62 & 16380 & 3.34 \textcolor{red}{(0.22$\uparrow$)} \\
& & \cellcolor{cellcolor}ChartIR (ours)  & \cellcolor{cellcolor}0.45 & \cellcolor{cellcolor}0.67 & \cellcolor{cellcolor}0.76 & \cellcolor{cellcolor}0.50 & \cellcolor{cellcolor}13.61 & \cellcolor{cellcolor}0.69 & \cellcolor{cellcolor}11995 & \cellcolor{cellcolor}{\textbf{{3.54 \textcolor{red}{(0.42$\uparrow$)}}}} \\
\cmidrule(lr){2-11}
&\multirow{3}{*}{GPT-4o} 
& Direct   & 0.70 & 0.80 & 0.89 & 0.74 & 13.53 & 0.68 & 12746 & 5.61 \\
&& METAL    & 0.83 & 0.86 & 0.86 & 0.71 & 12.50 & 0.66 & 13864 & 6.02 \textcolor{red}{(0.41$\uparrow$)} \\
&& \cellcolor{cellcolor}ChartIR (ours)  & \cellcolor{cellcolor}0.70 & \cellcolor{cellcolor}0.79 & \cellcolor{cellcolor}0.95 & \cellcolor{cellcolor}0.68 & \cellcolor{cellcolor}14.29 & \cellcolor{cellcolor}0.69 & \cellcolor{cellcolor}10676 & \cellcolor{cellcolor}{\textbf{{6.56 \textcolor{red}{(0.95$\uparrow$)}}}} \\
\midrule
\multirow{6}{*}{\rotatebox{90}{{ChartMimic}}}&\multirow{3}{*}{Qwen2-VL} 
& Direct   & 0.30 & 0.41 & 0.78 & 0.41 & 11.73 & 0.60 & 16644 & 2.20 \\
&& METAL    & 0.35 & 0.40 & 0.78 & 0.33 & 11.74 & 0.60 & 16461 &  2.32 \textcolor{red}{(0.12$\uparrow$)}    \\
&& \cellcolor{cellcolor}ChartIR (ours)  & \cellcolor{cellcolor}0.48 & \cellcolor{cellcolor}0.64 & \cellcolor{cellcolor}0.75 & \cellcolor{cellcolor}0.49 & \cellcolor{cellcolor}12.01 & \cellcolor{cellcolor}0.60 & \cellcolor{cellcolor}15127 & \cellcolor{cellcolor}{\textbf{{3.86 \textcolor{red}{(1.66$\uparrow$)}}}} \\
\cmidrule(lr){2-11}
&\multirow{3}{*}{GPT-4o} 
& Direct   & 0.78 & 0.82 & 0.97 & 0.74 & 13.05 & 0.64 & 11846 & 6.62 \\
&& METAL    & 0.86 & 0.84 & 0.96 & 0.80 & 13.09 & 0.70 & 11356 & 6.89 \textcolor{red}{(0.27$\uparrow$)} \\
&& \cellcolor{cellcolor}ChartIR (ours)  & \cellcolor{cellcolor}0.82 & \cellcolor{cellcolor}0.84 & \cellcolor{cellcolor}0.96 & \cellcolor{cellcolor}0.76 & \cellcolor{cellcolor}13.68 & \cellcolor{cellcolor}0.67 & \cellcolor{cellcolor}10129 & \cellcolor{cellcolor}{\textbf{{7.15 \textcolor{red}{(0.53$\uparrow$)}}}} \\

\bottomrule
\end{tabular}}
\end{table}

As shown in Table~\ref{tab:results}, the experimental results on the Plot2Code and ChartMimic datasets demonstrate that ChartIR consistently outperforms baseline methods across most evaluation metrics. These results validate our core objective, enhancing the chart-to-code capabilities of base multimodal models like GPT-4o and Qwen2-VL, and also show that ChartIR provides consistent improvements over both Direct Generation and existing approaches such as METAL.

\vspace{0.5em}
\noindent \textbf{GPT-4o Results}. For the GPT-4o model, ChartIR consistently outperforms both direct generation and METAL under all major evaluation metrics. On the Plot2Code dataset, direct generation and METAL achieve GPT-4o Scores of 5.61 and 6.02, respectively. In contrast, ChartIR reaches a score of 6.56, representing a relative improvement of 17\% over direct generation. Besides the overall score, ChartIR also improves across low-level metrics (Text, Type, Layout, Color) and traditional metrics (PSNR, SSIM, MSE), demonstrating better structural preservation and appearance similarity. On the ChartMimic dataset, ChartIR again shows the strongest performance, achieving a GPT-4o Score of 7.15, compared to 6.62 for direct generation and 6.89 for METAL. These results highlight ChartIR's superiority in generating chart code that best aligns with both structural accuracy and visual fidelity, particularly under the strong reasoning capabilities of GPT-4o.

\vspace{0.5em}
\noindent \textbf{Qwen2-VL Results}. For the open-source model Qwen2-VL, ChartIR also brings consistent and notable gains. On the Plot2Code dataset, GPT-4o Score improves from 3.12 (direct) and 3.34 (METAL) to 3.54 using ChartIR. Notably, in addition to higher structural accuracy (Layout +0.16), ChartIR greatly improves traditional metrics such as PSNR (13.61 vs. 11.91) and SSIM (0.69 vs. 0.62), highlighting enhanced image reconstruction quality. On the ChartMimic dataset, Qwen2-VL with ChartIR reaches a GPT-4o Score of 3.86, significantly outperforming both direct (2.20) and METAL (2.32). Low-level metrics also show marked gains (Text from 0.30 to 0.48, Type from 0.41 to 0.64), confirming the broad effectiveness of our approach in challenging visual conditions.

\vspace{0.5em}
\noindent \textbf{Analysis}. These results validate the strength of ChartIR in enhancing both open-source and closed-source multimodal models in the chart-to-code generation task. Compared to METAL, which also utilizes iterative refinement, our method achieves stronger layout alignment and semantic expressiveness. We believe this is due to the incorporation of an intermediate structured representation that retains chart semantics and layout cues while being model-friendly for generation. Furthermore, the improvements in the GPT-4o Score, a metric closely aligned with human preference, suggest that the benefits of ChartIR go beyond token-level or image-level matching. Instead, ChartIR enhances the holistic perception of the chart as a visually and semantically coherent artifact. This supports our motivation to incorporate intermediate representations that bridge the modality gap between vision and code, thereby benefiting both symbolic alignment and perceptual faithfulness.

\vspace{0.5em}
In summary, across both models and datasets, ChartIR leads to measurable and consistent improvements, highlighting its potential as a general-purpose enhancer for chart-to-code generation tasks. It performs robustly across diverse settings and model capabilities, making it a strong foundation for future multimodal reasoning and generation systems.

\subsection{Ablation Study}

To better understand the contribution of different components in ChartIR, we conduct an ablation study focusing on the two key elements: description and difference. Specifically, we randomly sample 20\% of examples from both the ChartMimic and Plot2Code datasets, resulting in a combined test set of 132 samples for evaluation.

We perform three separate experiments using both the open-source model Qwen2-VL and closed-source model GPT-4o on this ablation set:
    1) {ChartIR}: the complete version incorporating both the description and difference components;
    2) {Only Difference}: a variant where the description part is removed, and only the difference information is retained;
    and 3) {Only Description}: a variant where the difference information is removed, and only the description is used.

This setup allows us to isolate the impact of each component and evaluate how much each contributes to the model's ability to generate accurate chart code from images.

\paragraph{Effect of Description.}  
We first investigate the impact of description component in ChartIR. This component provides a natural language description of the visual content, offering high-level semantic guidance. As shown in Table~\ref{tab:ablation_results}, removing the description results in a noticeable performance drop. For Qwen2-VL, the GPT-4o score decreases from 3.96 to 3.10, and the average score of low-level structural metrics, \emph{Text}, \emph{Chart Type}, \emph{Layout}, and \emph{Color}, drops from 0.61 to 0.51. In addition, traditional similarity metrics also show consistent degradation. While for GPT-4o, the GPT-4o score decreases from 7.13 to 7.00. On the other metrics, ChartIR and Only Difference achieved comparable results. These results suggest that the description plays a critical role in helping the model better understand the chart’s overall structure and semantics.

\paragraph{Effect of Difference.}  
We then ablate the \textit{difference} component, which highlights the key variations between the current chart and its potential reference template. Without this explicit contrastive information, the model must infer visual and structural changes solely from the image, which increases the generation difficulty. The results demonstrate a similar trend: for Qwen2-VL, the GPT-4o score drops from 3.96 to 3.38, and the average of the four structural metrics decreases from 0.61 to 0.50; for GPT-4o, the GPT-4o score drops from 7.13 to 6.91, and the four low-level metrics also show a such decline. Furthermore, traditional similarity-based measures decline in both Qwen2-VL and GPT-4o, confirming that explicitly modeling chart-specific differences significantly improves generation accuracy, particularly in compositional reasoning scenarios.

\begin{table*}[t]
\caption{Ablation study results on ChartIR components. We compare full ChartIR with variants using only static descriptions or difference-based prompts.}
\label{tab:ablation_results}
\centering
\vspace{2mm}
\small
\renewcommand{\arraystretch}{1.3}
\setlength{\tabcolsep}{6pt}
\resizebox{0.95\linewidth}{!}{
\begin{tabular}{llccccccc|c}
\toprule
\multirow{2}{*}{\textbf{Base Model}} & \multirow{2}{*}{{\textbf{Experiment Setting}}} & \multicolumn{4}{c}{\textbf{Low-Level Metrics}} & \multicolumn{3}{c|}{\textbf{Traditional Metrics}} & \multirow{2}{*}{\textbf{GPT-4o Score}} \\
\cmidrule(lr){3-6} \cmidrule(lr){7-9}
 & & \textbf{Text} & \textbf{Type} & \textbf{Layout} & \textbf{Color} & \textbf{PSNR} & \textbf{SSIM} & \textbf{MSE} & \\
\midrule
\multirow{3}{*}{Qwen2-VL}
& \cellcolor{cellcolor}ChartIR (Both Components)             & \cellcolor{cellcolor}0.42 & \cellcolor{cellcolor}0.66 &\cellcolor{cellcolor} 0.81 & \cellcolor{cellcolor}0.54  &\cellcolor{cellcolor} 12.30 &\cellcolor{cellcolor} 0.65 &\cellcolor{cellcolor} 14500 & \cellcolor{cellcolor}3.96 \\
& \hspace{1em} $\triangleright$ \emph{Only Description}    & 0.35 & 0.54 & 0.68 & 0.43  & 12.47 & 0.63 & 14299 & 3.38 \textcolor{green!70!black}{\textbf{(0.58$\downarrow$)}} \\
& \hspace{1em} $\triangleright$ \emph{Only Difference}     & 0.37 & 0.53 & 0.69 & 0.45  & 12.26 & 0.64 & 15070 & 3.10 \textcolor{green!70!black}{\textbf{(0.86$\downarrow$)}} \\
\midrule
\multirow{3}{*}{GPT-4o}
 &\cellcolor{cellcolor} ChartIR  (Both Components)           & \cellcolor{cellcolor}0.80 & \cellcolor{cellcolor}0.81 & \cellcolor{cellcolor}0.69 & \cellcolor{cellcolor}0.94 & \cellcolor{cellcolor}14.09 & \cellcolor{cellcolor}0.70 & \cellcolor{cellcolor}9144 & \cellcolor{cellcolor}7.13 \\
& \hspace{1em} $\triangleright$ \emph{Only Description}    & 0.76 & 0.81 & 0.69 & 0.92 & 13.76 & 0.69 & 9883 & 6.91 \textcolor{green!70!black}{\textbf{(0.22$\downarrow$)}} \\
& \hspace{1em} $\triangleright$ \emph{Only Difference}   & 0.79 & 0.83 & 0.71 & 0.91 & 14.14 & 0.70 & 8976 & 7.00 \textcolor{green!70!black}{\textbf{(0.13$\downarrow$)}} \\
\bottomrule
\end{tabular}}
\end{table*}

In summary, the ablation study clearly demonstrates that both description and difference components are essential to ChartIR, with their combined use yielding the best results across all metrics. While the description provides crucial semantic context for chart understanding, the difference component enables precise identification of structural variations, and their synergistic effect is particularly pronounced for open-source models like Qwen2-VL. These findings validate our design choice of incorporating both complementary representations in ChartIR's intermediate format.

\section{Discussion and conclusion}

\textbf{Limitations.} Although ChartIR achieves superior performance compared to direct prompting and Metal, it exhibits two limitations: 1) Compared to METAL, our method requires more computational cost, but since it makes fewer queries, it results in lower time overhead on closed-source models. 2) Limited effectiveness on closed-source models: While our structured instruction demonstrates strong capabilities on open-source models, its impact on closed-source models is relatively limited. This is mainly because closed-source models like GPT-4o possess strong chart understanding abilities, rendering structured instructions less effective. We believe that the performance of ChartIR on such models could be further enhanced with a more powerful instruction generation model.

\textbf{Conclusion.} In this work, we present ChartIR, a training-free and model-agnostic framework designed to enhance chart-to-code generation in MLLMs. Unlike prior approaches that rely on either direct prompting or metric-driven refinement, ChartIR introduces a structured, two-stage process consisting of multi-dimensional chart description and iterative difference-based refinement. This design directly targets the core challenges of chart understanding and faithful code translation. Through comprehensive experiments on two benchmark datasets, Plot2Code and ChartMimic, we demonstrate that ChartIR significantly improves over both direct generation and METAL, across a broad range of evaluation metrics. Notably, our ablation studies validate the essential roles of both the chart description and difference. ChartIR’s generality and compatibility with both open-source and commercial MLLMs suggest that it can serve as a robust and extensible solution for future work in multimodal chart understanding and generation.

\section{Acknowledgments}

This project is supported by National Natural Science Foundation of China (No.\ 62406192), Shanghai Municipal Special Program for Basic Research on General AI Foundation Models (Grant No. 2025SHZDZX025G03), Tencent WeChat Rhino-Bird Focused Research Program, and Kuaishou Technology.

\bibliography{iclr2026_conference}
\bibliographystyle{iclr2026_conference}

\newpage

\begin{center}
    \Large \textbf{Appendix}\\[0.5cm]
\end{center}
\subsection*{A Prompt Template}

In this appendix, we present the prompt templates used in our experiments. 

\subsubsection*{A.1 Prompt for Initial Code Generation}

The following prompt is used to guide the model in extracting structural and stylistic information from a chart image. It asks the model to describe the chart’s key visual elements without generating any code. 

\begin{tcolorbox}
Please analyze the provided image, which was generated using Python's matplotlib.pyplot. Your task is to identify and describe the key visual elements and details necessary to recreate a similar figure. Focus on the following aspects:

1. The number and location of subplots: If the chart consists of multiple subplots, please describe the number of subplots. Meanwhile, specify the function (such as subplot(212), subplot(221), subplot(311), etc.) that should be used for each subplot to place it correctly in the figure. Otherwise, please answer "The chart contains only one subplot."

2. Plot type: Identify the type of the plot, such as line plot, pie chart, scatter plot, bar chart, histogram, event plot, node plot, radar plot, area plot, box plot, heatmap, bubble chart, polar plot, violin plot, text plot, etc. If the chart does not fit any known chart type, please describe which matplotlib functions should be used to generate the visual elements.

3. Axes: Describe the labels, titles, and scales (e.g., linear, logarithmic) of the x-axis and y-axis.

4. Color: Describe the color schemes. If the color is a fixed value, please specify the exact color or name; if the color is dynamically generated, please describe the generation logic, such as random generation, data mapping, algorithmic calculation, or the use of a specific colormap.

5. Styles: According to the type of chart, describe the stylistic features that should be noted, such as orientation, width, size, and marker types.

6. Annotations and Texts: Mention any text annotations, labels, titles, or legends present in the image.

7. Grid and Background: Describe whether a grid is present and any background elements.

8. Data Characteristics: If the data is generated randomly, specify the distribution used (e.g., normal, uniform, exponential, Poisson, binomial, geometric, gamma, beta, etc.). If the chart is generated from a function, identify the function type (e.g., sine, exponential growth, mixed type, etc.). For periodic functions, also include the period and phase information. If the chart uses specific given values, estimate the approximate magnitude of the data.

Note: If the chart contains multiple subplots, you must describe all the above information for each subplot. You do not need to generate any code.
\end{tcolorbox}

Once the description is obtained, the following prompt is used to generate the corresponding Python code based on the extracted details.

\begin{tcolorbox}
Here's a description of the image: $<$\emph{description}$>$. According to the description, please generate the Python script used to draw this image.
\end{tcolorbox}

\subsubsection*{A.2 Prompt for Iterative Refinement}

In this stage, we aim to refine previously generated code by capturing the differences between two chart images. The following prompt is used to guide the model in identifying meaningful visual differences between two images.

\begin{tcolorbox}
Please compare the two provided images generated using Python code. Your task is to describe the differences between the first image and the second image. Focus on the following aspects:

1. Axes: Describe the differences in the labels, titles, and scales (e.g., linear, logarithmic) of the x-axis and y-axis between the two images.

2. Color: Describe the differences in the color schemes of the two images.

3. Styles: According to the type of chart, describe the differences in styles that should be noted in the charts, such as orientation, width, size, marker types, etc.

4. Annotations and Texts: Mention any differences in text annotations, labels, titles, or legends present in the two images.

5. Grid and Background: Describe any differences in the presence of a grid and background elements between the two images.

6. Others: Include any other significant differences not covered by the above aspects.

Note: If the charts contain multiple subplots, you need to describe the differences for each subplot.
\end{tcolorbox}

Given the visual differences and the original code, the following prompt is used to instruct the model to modify the existing code to match the second image.

\begin{tcolorbox}
Here is the Python script used to generate the first image:$<$\emph{code}$>$

I want to modify the script to draw the second image.

The difference between the two images generated by the given code is: $<$\emph{difference}$>$

The description of the second image is: $<$\emph{description}$>$

Based on the differences and the description, please provide the Python code that can be used to generate the second image.
\end{tcolorbox}

\subsubsection*{A.3 Prompt for Evaluation}

The following prompt is used to instruct GPT-4o to assess the visual similarity between the ground truth image and the image generated from model-produced code.

\begin{tcolorbox}
You are a helpful assistant. Please evaluate the similarity between a reference image created using matplotlib and an image generated by code provided by an AI assistant. Consider factors such as the overall appearance, colors, shapes, positions, and other visual elements of the images. Begin your evaluation by providing a short explanation. Be as objective as possible.

After providing your explanation, you must rate the response on a scale of 1 to 10 by strictly following this format: \texttt{"Rating: [[5]]"}

The reference image is the first image and the generated image is the second image.
\end{tcolorbox}

\subsection*{A.4 Prompt for ChartX Code Description Generation}

The following prompt is designed for instructing GPT-4o to analyze a given Python script and generate a detailed textual description of the resulting chart. This description focuses on the visual and stylistic elements of the figure, enabling its recreation without direct access to the code.

\begin{tcolorbox}
    Please analyze the provided code, which used Python's matplotlib.pyplot to generate an image. Your task is to identify and describe the key visual elements and details necessary to recreate a similar figure. Focus on the following aspects:
    
    1. Plot Type: Identify the type of plot (e.g., line plot, scatter plot, bar chart, histogram, etc.).
    
    2. Axes: Describe the labels, titles, and scales (e.g., linear, logarithmic) of the x-axis and y-axis.
    
    3. Data Representation: Summarize how the data is represented (e.g., points, lines, bars, colors, markers, etc.).
    
    4. Colors and Styles: Note the color schemes, line styles, marker shapes, or other stylistic elements used.
    
    5. Annotations and Texts: Mention any text annotations, labels, titles, or legends present in the image.
    
    6. Grid and Background: Describe whether a grid is present and any background elements.
    
    7. Data Characteristics: If the data is visible, describe its general trend, distribution, or any notable patterns (e.g., clusters, peaks, trends).
    
    8. Random Seed or Data Source: If the image involves randomness, extract the random seed or describe the data source for reproducibility.
    
    Use clear, everyday language to describe the key elements of the code. Your response should provide enough information for someone to redraw the figure without needing to see the original code.Here is the code:$<$\emph{code}$>$
\end{tcolorbox}

\subsection*{B Detailed training settings}
We trained Qwen2.5-VL for generating descriptions on 4 A100 GPUs for approximately one day. We include the detailed hyperparameters for training Qwen2.5-VL in Table~\ref{tab:training_params}.
\begin{table}[htbp]
\centering
\caption{Training hyperparameters}
\label{tab:training_params}
\vspace{2mm}
\resizebox{0.95\linewidth}{!}{
\begin{tabular}{cccccccc}
\toprule
\textbf{Epochs} & \textbf{Lora\_rank} &\textbf{ViT freeze}& \textbf{Learning rate} & \textbf{Warmup\_ratio} & \textbf{Numerical precision} & \textbf{Scheduler} \\
\midrule
30 & 64 & yes & $5.0e^{-5}$ & 0.03 & bfloat16 & cosine \\
\bottomrule
\end{tabular}}
\end{table}

\subsection*{C Broader Impact}

This work investigates the task of chart-to-code generation, which aims to translate visual chart representations into their corresponding code. Such technology holds promise for enhancing accessibility and efficiency in data analysis workflows. For example, it can support users with limited programming expertise in reproducing and modifying visualizations, and enable the reverse engineering of figures from educational or scientific publications. Nonetheless, the automatic generation process may introduce errors or misleading outputs, particularly when the source charts contain design flaws or follow non-standard conventions.

\subsection*{D Case study}
\begin{figure}[ht]
    \centering
    \includegraphics[width=1\textwidth]{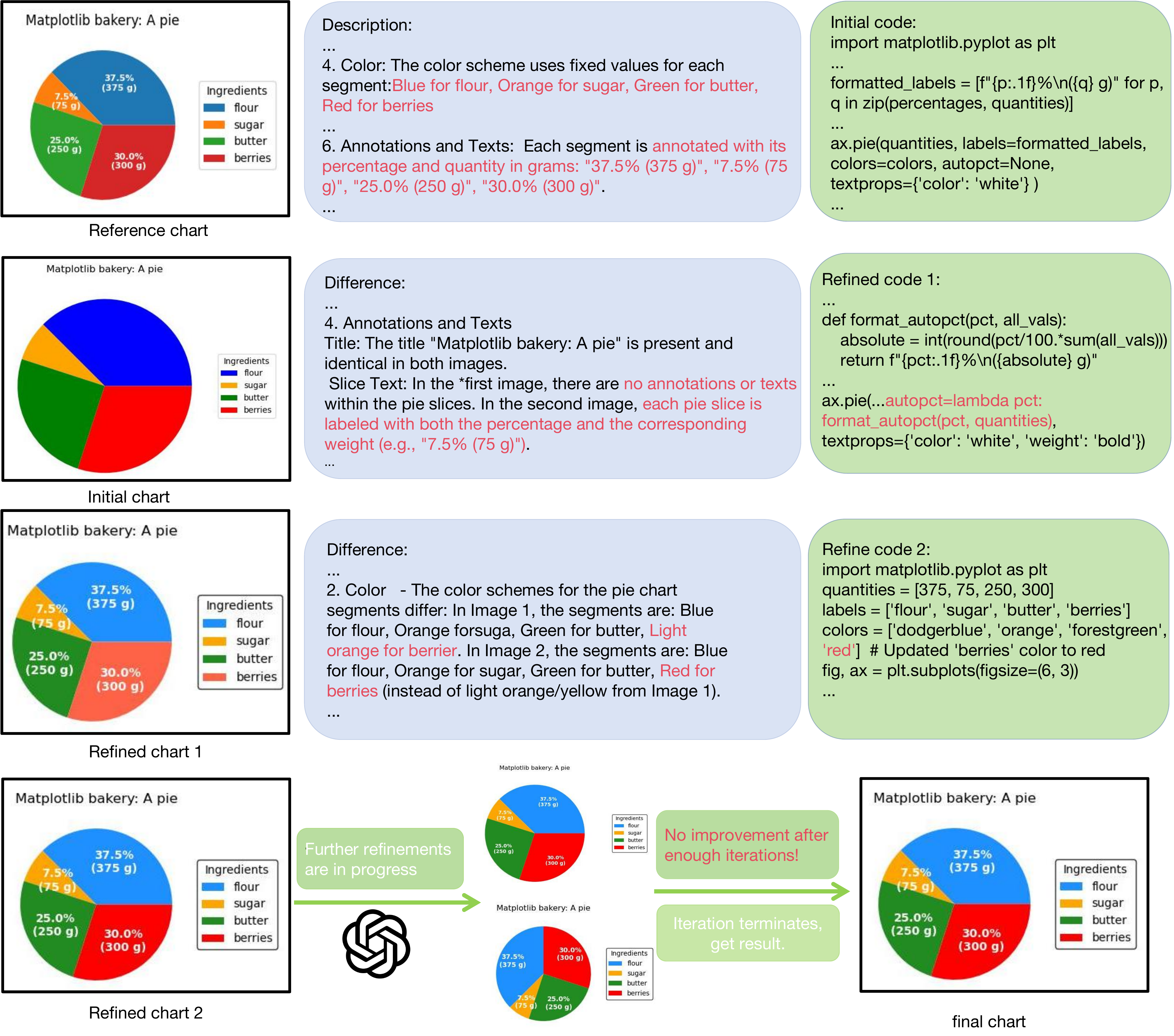} 
    \caption{A case study based on {ChartIR}.The initial chart had missing textual information. The first refinement recovered the text but caused a color error, while the second refinement fixed the color error without new issues, resulting in a chart with better overall performance than the initial version.}\label{fig:demo}
\end{figure}

A case study based on {ChartIR} (Figure~\ref{fig:demo}): in the initial code generation stage, a description of the reference chart is obtained to generate initial code that renders correct segment colors but missing textual annotations; in the first refinement stage, the missing annotations are identified and fixed, but a segment color error is introduced; in the second refinement stage, the color error is corrected without new issues, and subsequent iterations fail to improve further, so the refined chart 2 is selected as the final output with more accurate annotations and colors than the initial chart.

\subsection*{E Examples}
\begin{figure*}[p]
\centering
\begin{subfigure}{0.45\textwidth}
    \centering
    \includegraphics[width=\linewidth]{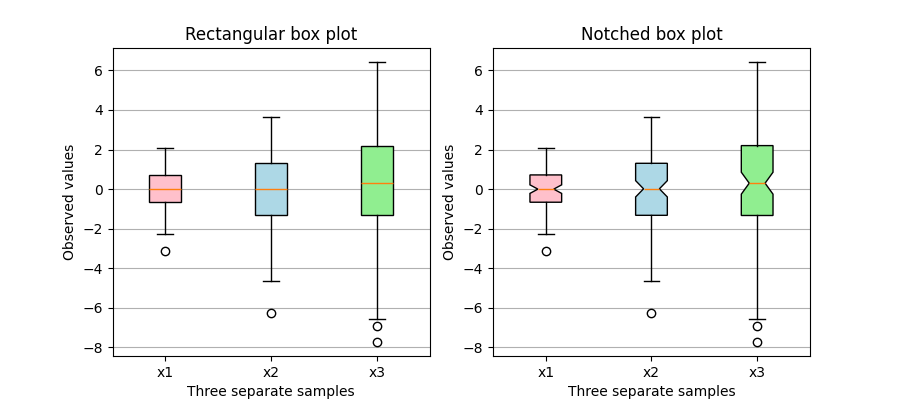}
    \caption{Ground Truth}
\end{subfigure}
\hfill
\begin{subfigure}{0.45\textwidth}
    \centering
    \includegraphics[width=\linewidth]{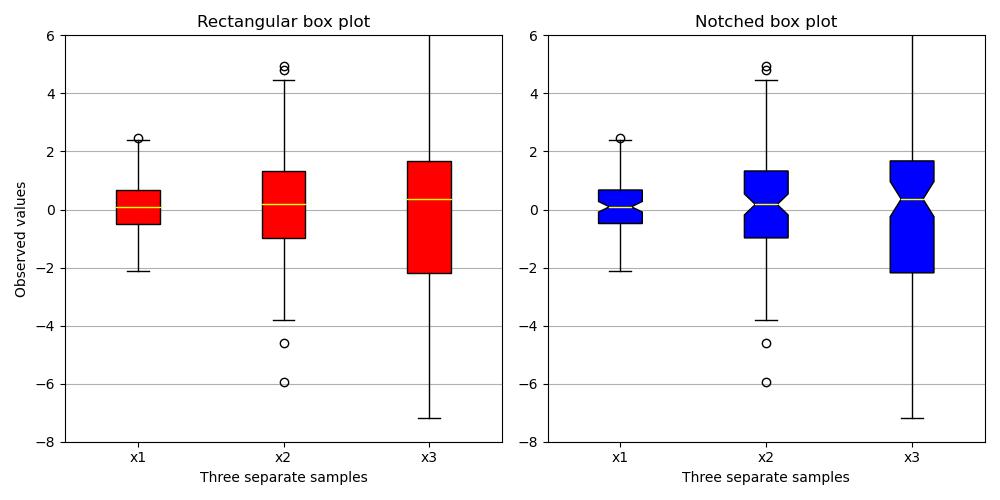}
    \caption{Direct}
\end{subfigure}
\vspace{0.2cm}
\begin{subfigure}{0.45\textwidth}
    \centering
    \includegraphics[width=\linewidth]{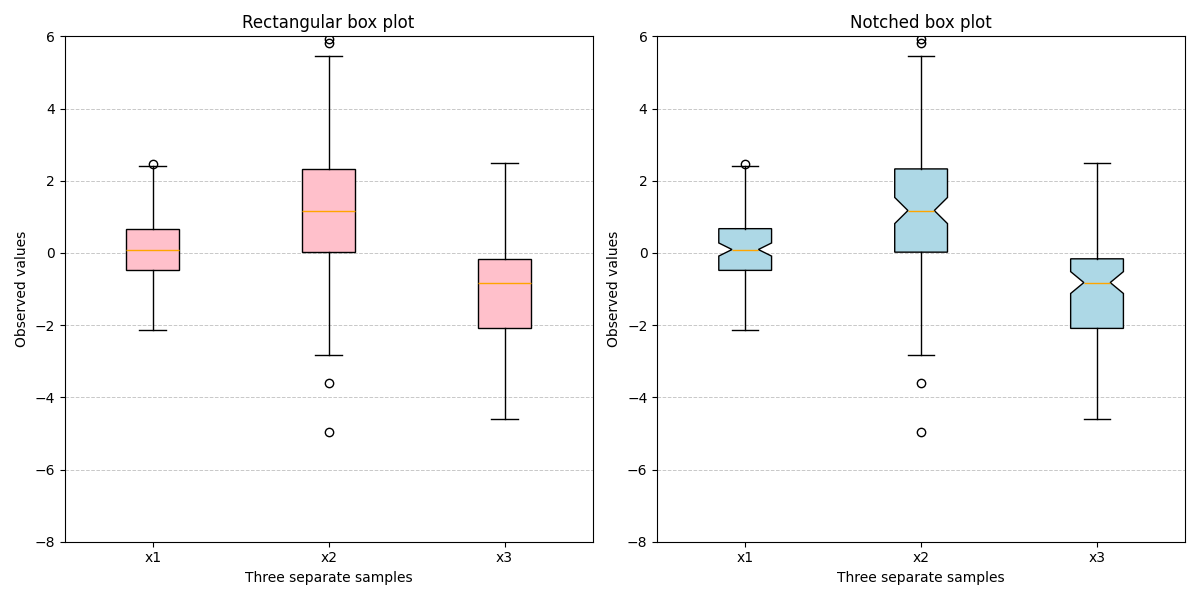}
    \caption{METAL}
\end{subfigure}
\hfill
\begin{subfigure}{0.45\textwidth}
    \centering
    \includegraphics[width=\linewidth]{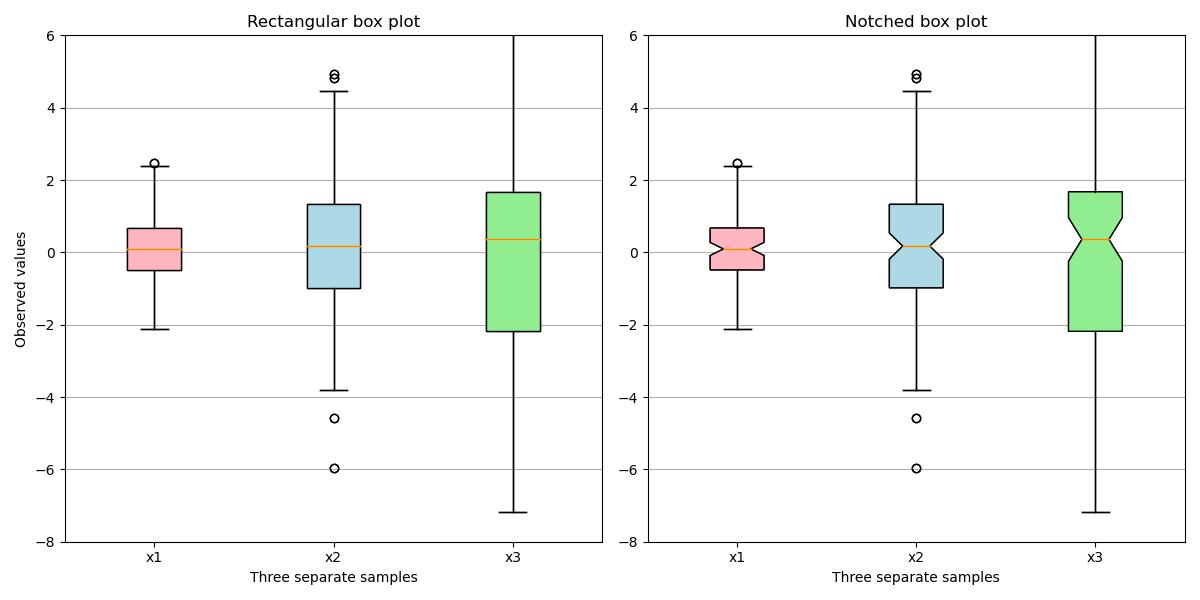}
    \caption{ChartIR}
\end{subfigure}
\caption{Visual comparison of chart generation results across different methods (Example 1).}
\label{fig:example1}
\end{figure*}

\begin{figure*}[p]
\centering
\begin{subfigure}{0.45\textwidth}
    \centering
    \includegraphics[width=\linewidth]{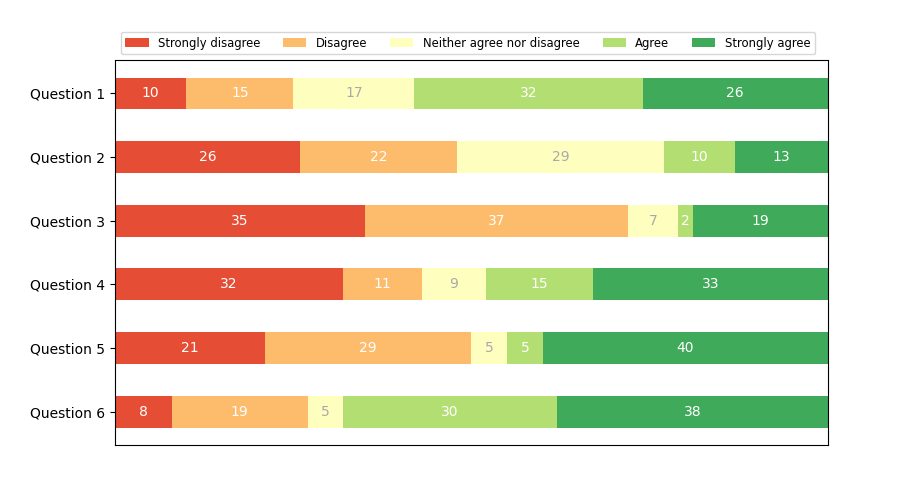}
    \caption{Ground Truth}
\end{subfigure}
\hfill
\begin{subfigure}{0.45\textwidth}
    \centering
    \includegraphics[width=\linewidth]{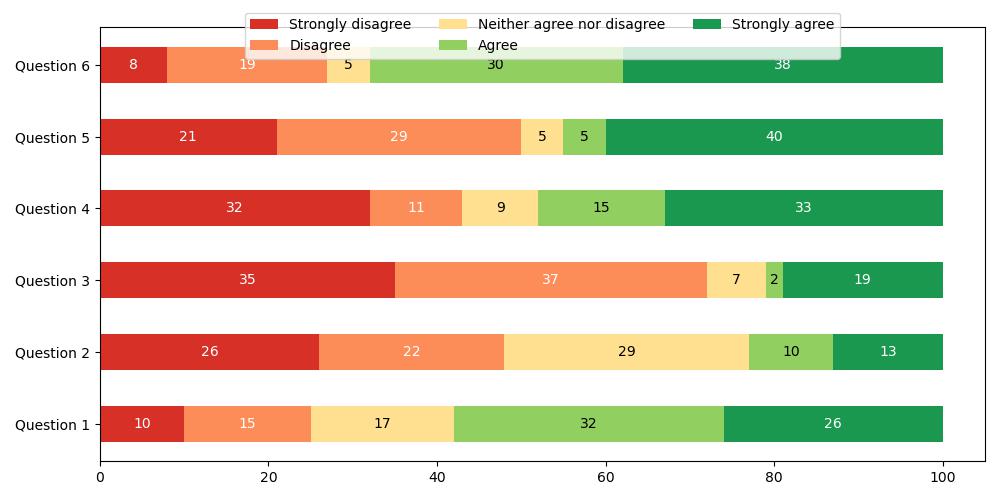}
    \caption{Direct}
\end{subfigure}
\vspace{0.2cm}
\begin{subfigure}{0.45\textwidth}
    \centering
    \includegraphics[width=\linewidth]{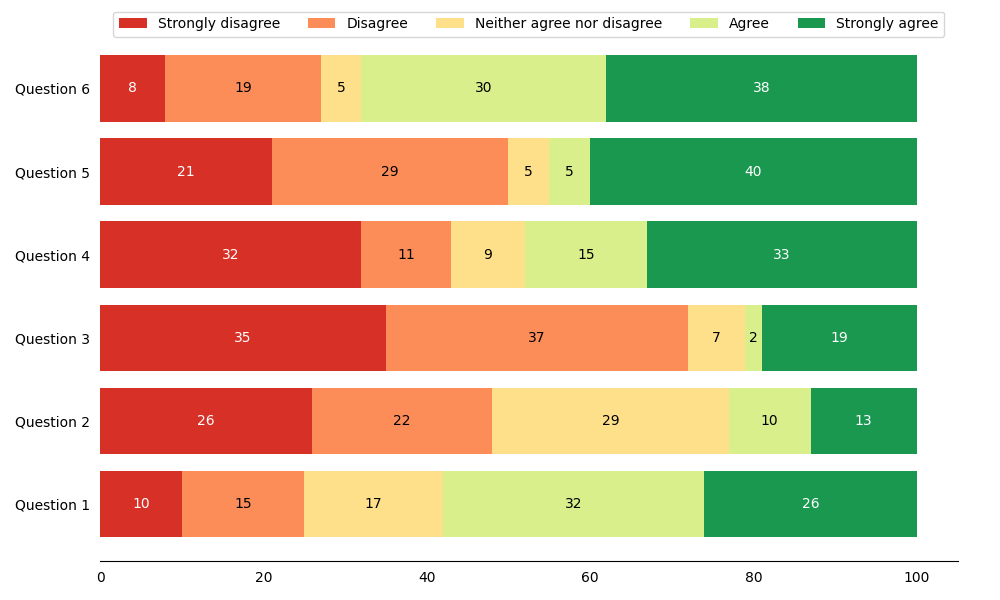}
    \caption{METAL}
\end{subfigure}
\hfill
\begin{subfigure}{0.45\textwidth}
    \centering
    \includegraphics[width=\linewidth]{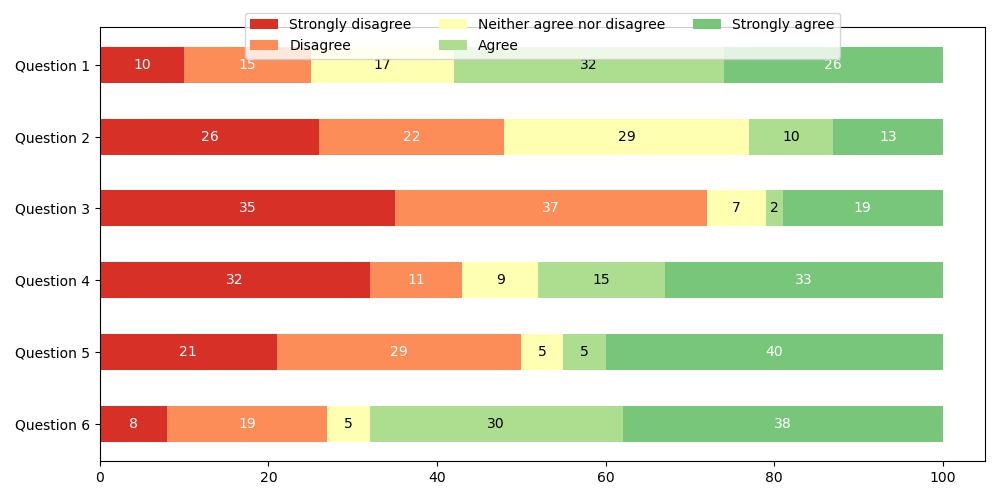}
    \caption{ChartIR}
\end{subfigure}
\caption{Visual comparison of chart generation results across different methods (Example 2).}
\label{fig:example2}
\end{figure*}

\begin{figure*}[p]
\centering
\begin{subfigure}{0.45\textwidth}
    \centering
    \includegraphics[width=\linewidth]{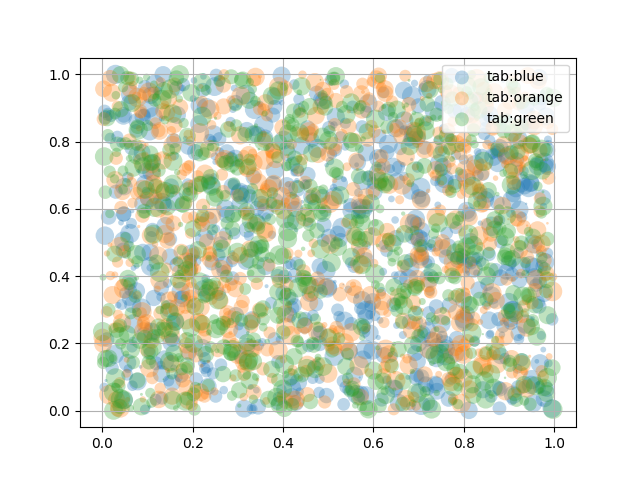}
    \caption{Ground Truth}
\end{subfigure}
\hfill
\begin{subfigure}{0.45\textwidth}
    \centering
    \includegraphics[width=\linewidth]{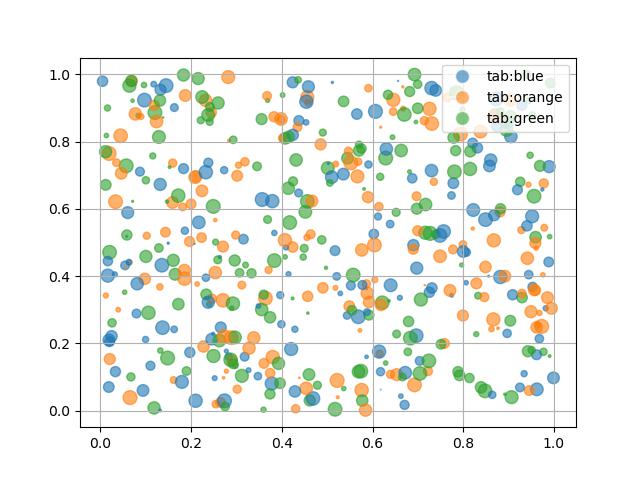}
    \caption{Direct}
\end{subfigure}
\vspace{0.2cm}
\begin{subfigure}{0.45\textwidth}
    \centering
    \includegraphics[width=\linewidth]{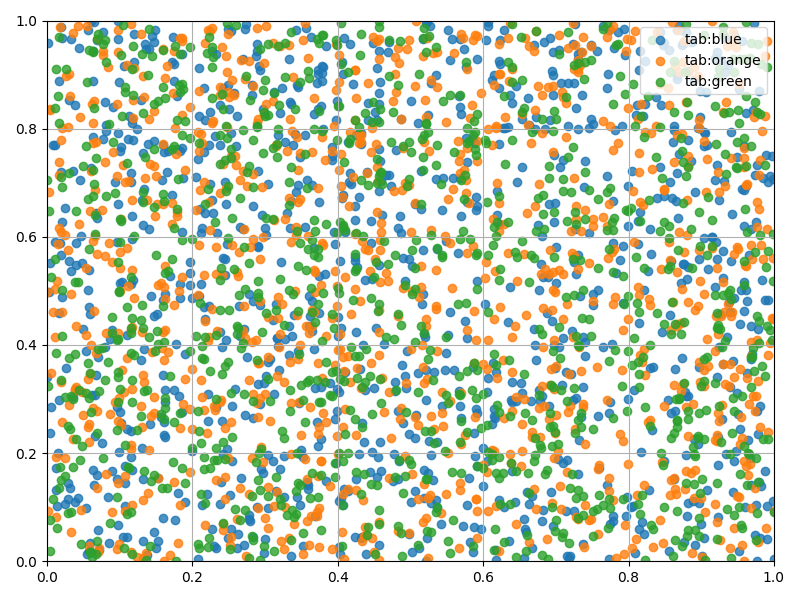}
    \caption{METAL}
\end{subfigure}
\hfill
\begin{subfigure}{0.45\textwidth}
    \centering
    \includegraphics[width=\linewidth]{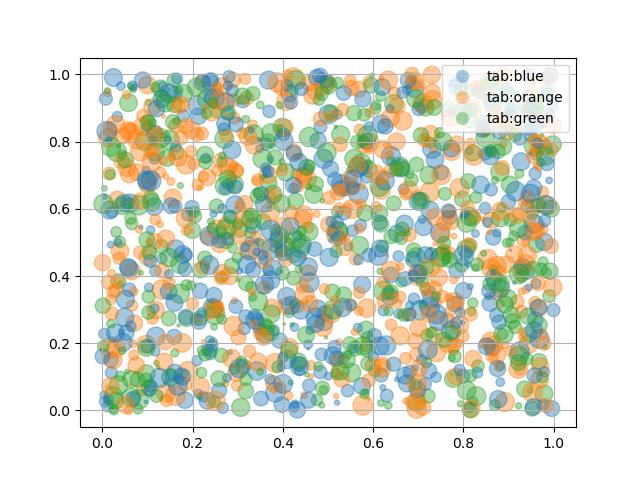}
    \caption{ChartIR}
\end{subfigure}
\caption{Visual comparison of chart generation results across different methods (Example 3).}
\label{fig:example3}
\end{figure*}

\begin{figure*}[p]
\centering
\begin{subfigure}{0.45\textwidth}
    \centering
    \includegraphics[width=\linewidth]{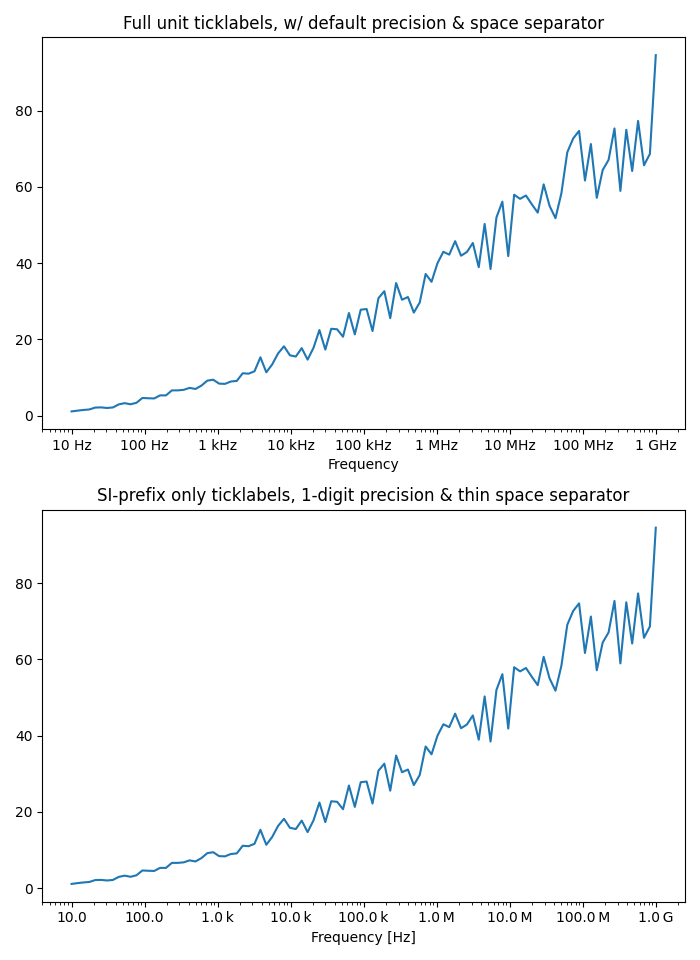}
    \caption{Ground Truth}
\end{subfigure}
\hfill
\begin{subfigure}{0.45\textwidth}
    \centering
    \includegraphics[width=\linewidth]{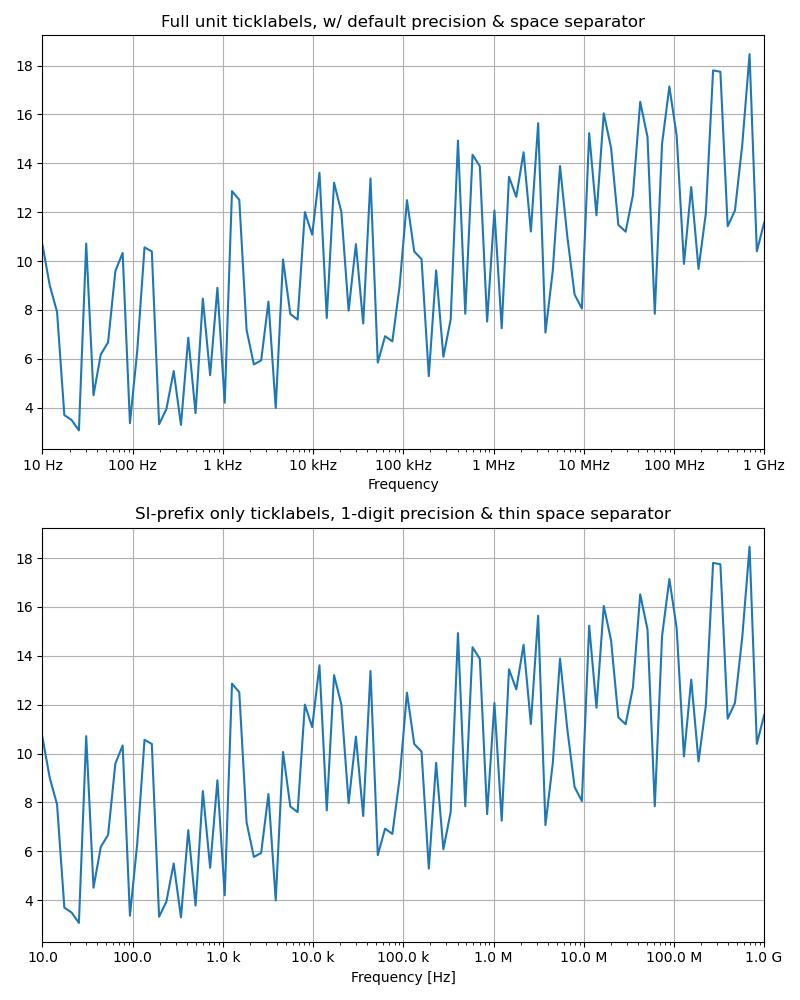}
    \caption{Direct}
\end{subfigure}
\vspace{0.2cm}
\begin{subfigure}{0.45\textwidth}
    \centering
    \includegraphics[width=\linewidth]{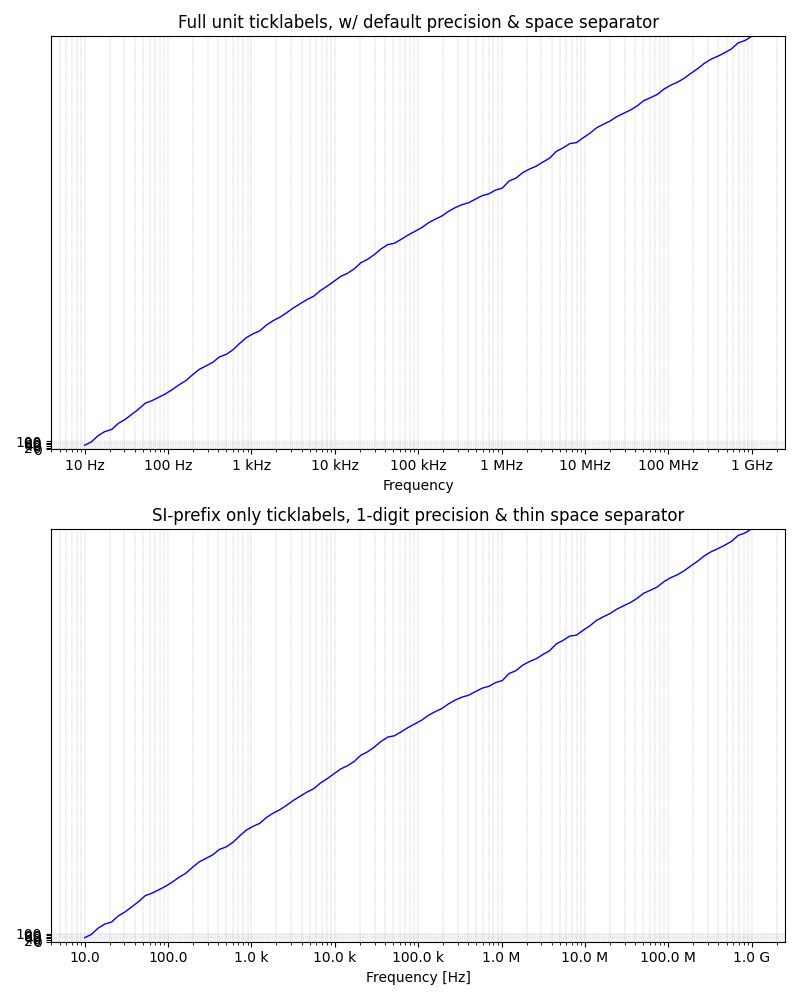}
    \caption{METAL}
\end{subfigure}
\hfill
\begin{subfigure}{0.45\textwidth}
    \centering
    \includegraphics[width=\linewidth]{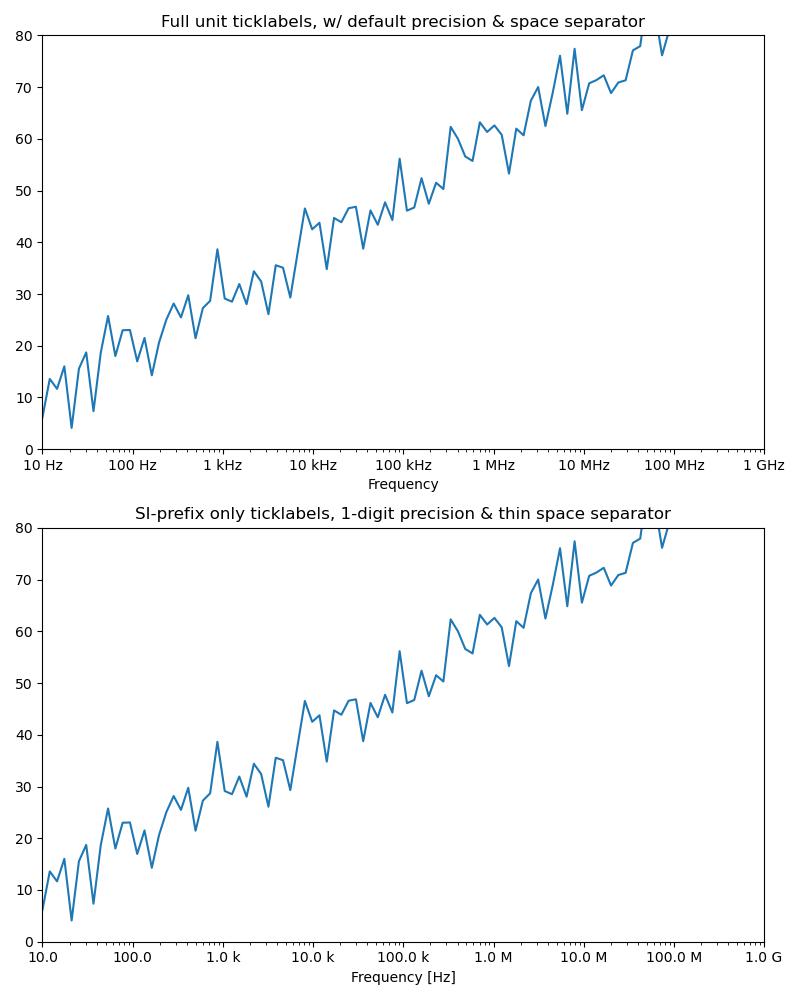}
    \caption{ChartIR}
\end{subfigure}
\caption{Visual comparison of chart generation results across different methods (Example 4).}
\label{fig:example4}
\end{figure*}

We provide several visual comparison examples to illustrate the differences among Direct, METAL, and ChartIR (Figures \ref{fig:example1}, \ref{fig:example2}, \ref{fig:example3}, \ref{fig:example4}) . For each example, we present the chart generated by each method alongside the ground truth chart, enabling a direct visual comparison of their generation quality.

\end{document}